\documentclass{article}
\usepackage{spconf,amsmath,graphicx,hyperref}
\usepackage{times}
\usepackage{latexsym}
\usepackage{multirow}
\usepackage[T1]{fontenc}

\usepackage[utf8]{inputenc}
\usepackage{listings}
\usepackage{xcolor}
\usepackage[table]{xcolor}
\definecolor{codegreen}{rgb}{0,0.6,0}
\definecolor{codegray}{rgb}{0.5,0.5,0.5}
\definecolor{codepurple}{rgb}{0.58,0,0.82}
\definecolor{backcolour}{rgb}{0.95,0.95,0.92}
\lstdefinestyle{mystyle}{
  columns=flexible,
  backgroundcolor=\color{backcolour}, commentstyle=\color{codegreen},
  keywordstyle=\color{black},
  numberstyle=\tiny\color{codegray},
  stringstyle=\color{codepurple},
  basicstyle=\ttfamily\scriptsize,
  breakatwhitespace=false,         
  breaklines=true,                 
  captionpos=b,                    
  keepspaces=true,                 
  numbers=left,                    
  numbersep=5pt,                  
  showspaces=false,                
  showstringspaces=false,
  showtabs=false,                  
  tabsize=2,
}
\title{Fine-Grained Customized Fashion Design with Image-into-Prompt benchmark and dataset from LMM}
\name{Hui Li\textsuperscript{1}, Yi You\textsuperscript{1}, Qiqi Chen\textsuperscript{1}, Bingfeng Zhang\textsuperscript{2}, George Q. Huang\textsuperscript{1}$^*$}
\address{\textsuperscript{1}The Hong Kong Polytechnic University, China\\
    \textsuperscript{2}China University of Petroleum (East China), China}

\begin{document}
%
\maketitle
\begin{abstract}
Generative AI evolves the execution of complex workflows in industry, where the large multimodal model empowers fashion design in the garment industry. Current generation AI models magically transform brainstorming into fancy designs easily, but the fine-grained customization still suffers from text uncertainty without professional background knowledge from end-users. Thus, we propose the Better Understanding Generation (BUG) workflow with LMM to automatically create and fine-grain customize the cloth designs from chat with image-into-prompt. Our framework unleashes users' creative potential beyond words and also lowers the barriers of clothing design/editing without further human involvement. To prove the effectiveness of our model, we propose a new FashionEdit dataset that simulates the real-world clothing design workflow, evaluated from generation similarity, user satisfaction, and quality. The code and dataset: https://github.com/detectiveli/FashionEdit.
\end{abstract}
\begin{keywords}
Image Editing, LMM, Fashion
\end{keywords}

\section{Introduction}
\label{sec:intro}

\begin{figure}[htb]
\centering
\includegraphics[width=0.8\linewidth]{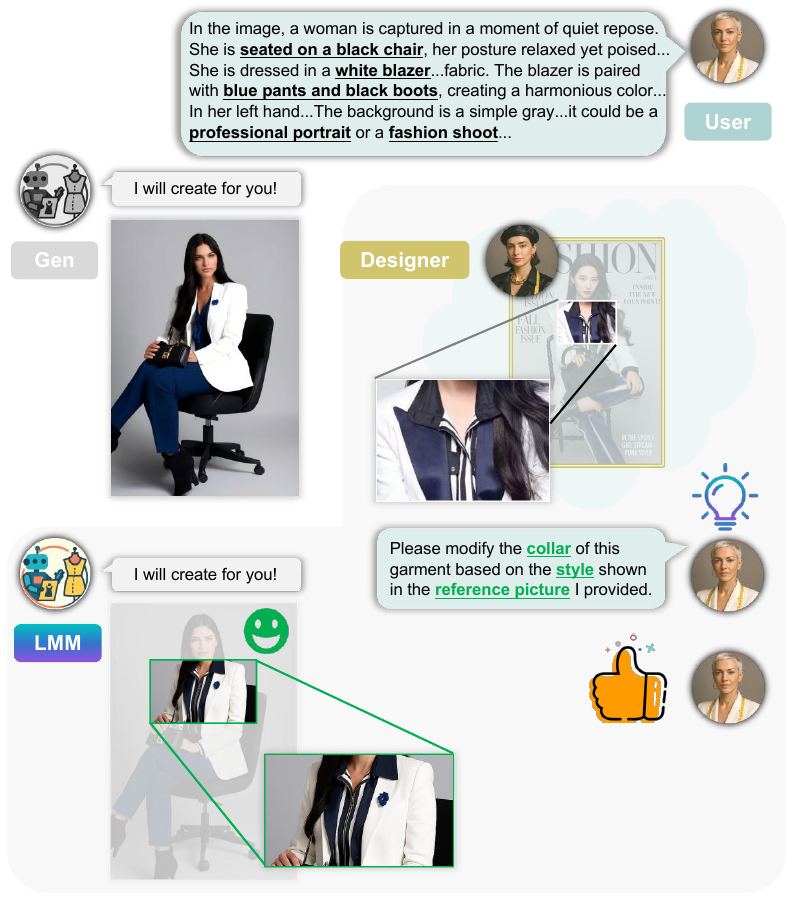}
\caption{Example of the real-world fine-grained customization fashion design.}
\label{fig:intro}
\end{figure}

\begin{figure*}[htb]
  \centering
  \includegraphics[width=\linewidth]{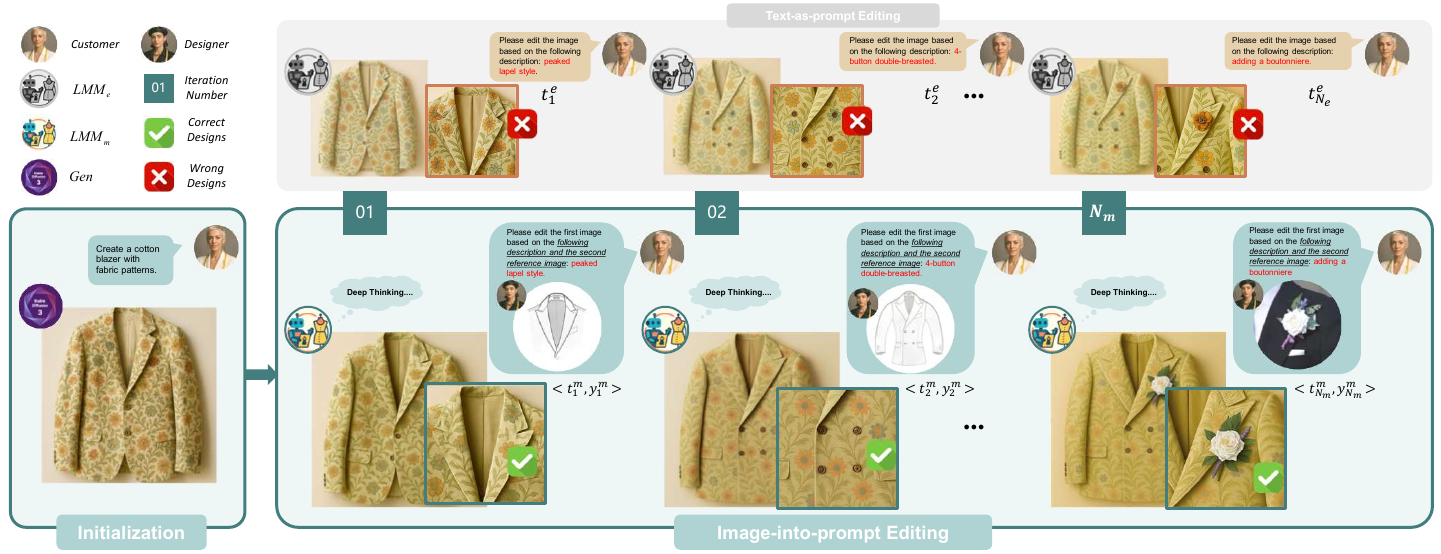}
  \caption{The workflow example of fine-grained customized fashion design. (1) \textbf{Initialization} generates a draft design from the user request based on SD3~\cite{esser2024scaling}. (2) \textbf{Image Editing} refine the design iteratively via text-as-prompt/image-into-prompt editing, with GPT4.1-mini~\cite{gpt41mini}. The three iteration results are modified referring to the sketch, the details, and the example image.}
  \label{fig:method}
\end{figure*}

Generative AI (GenAI) aims to execute complex workflows for humans. As one of the key components in GenAI, the development of the Large Multimodal Models (LMMs) enable new capabilities of AI agents in industry workflows, such as financial analysis \cite{nygaard2024news}, industrial solutions \cite{cheng2024shield}, specialized assistants \cite{kim2024mild}, and fashion design \cite{zhang2023diffcloth}, which are attributed to their rich understanding and execution ability.

In the garment industry, an order begins with customer needs, then goes through the designer, pattern maker~\cite{korosteleva2024garmentcodedata}, tailor~\cite{zhu2020deep,zhou2023clothesnet}, and finally ends with delivery~\cite{shi2025generative}. Current LMMs mainly focus on analyzing customers' needs to recommend items as their preferences \cite{liu2024sequential}. However, with the growing demand for personalized clothing, the customer is also willing to be their own designer, who creates and adjusts the design until satisfaction.

AI-generated fashion design focuses on customization based on natural language description (\textit{e.g.}, Stable Diffusion 3~\cite{esser2024scaling}, DALL-E 2~\cite{ramesh2022hierarchical}, which easily transform the sparklings into visual demonstrations~\cite{zhang2024fabricdiffusion}. However, pure text description struggles with ambiguity, as shown in Fig.\ref{fig:intro}, the description ``white blazer'' omits a detailed collar style that normally comes from a designer's professional skill. This raises the challenge: \textit{How to instruct the fashion generation to understand customer desires beyond simple description?}


In real-world scenarios, the customer shows a sample image as reference (\emph{e.g.}, an existing design from a fashion magazine) where the designer translates the principle into precise fashion elements. This inspires us to propose a new benchmark: \textit{Better Understanding Generation (BUG)} by showing AI the Image-into-Prompt, to meet the request of fine-grained customization in fashion design. BUG initializes a draft design image first, then continuously modifies the image not only following the user's text-prompts but also referring to image-prompts. Different from previous LMM image editing, our approach is more challenging than editing from real image~\cite{brooks2023instructpix2pix,zhang2024hive}, on object level modification~\cite{zhang2023magicbrush,wang2023imagen} or needing further human involvement~\cite{laput2013pixeltone}.

To evaluate the performance of this task, we propose a new FashionEdit dataset modified from DeepFashion-MultiModal~\cite{jiang2022text2human}. FashionEdit uses LMM to analyze two fashion fine-grained components, including the generated images and the differences between the generated and ground-truth images. The differences comprise the descriptions and cropped regions from the original images, corresponding to the user's desires and the referring images. We evaluate the performance of models using BUG on this dataset from content similarity (CLIP~\cite{radford2021learning}), user satisfaction (our CLIP*), and quality (PSNR). The CLIP* score increased 20.3\% between pure text and our BUG after three modifications, proving the effectiveness of our method.

\section{Methodology}

\subsection{Initialization}
\label{sec:IG}

The initialization of our benchmark uses a standard text-based image generation model (\emph{e.g.} SD3~\cite{esser2024scaling}), which inputs a fuzzy text generation-prompt $t^g$ and generates a design image $y^g$ defined as: $y^g = \mathit{Gen}(t^g).$

\subsection{Image Editing}
\label{sec:IEE}
\subsubsection{Text-as-prompt Editing}
\label{sec:TE}
The vanilla text-based image editing takes the initial design from Sec.\ref{sec:IG}, following an $N_e$-rounds of text editing-prompt $T^{e}$=$\{t_{i}^{e}\}^{N_e}_1$ from user to change the design. Correspondingly, the edited images are defined as $Y^{e}$=$\{y^{e}_i\}^{N_e}_1$ and each $y^{e}_i$ is generated by an editing $\mathit{LMM}_e$:

\begin{equation}
y^{e}_i = \mathit{LMM}_{e}(y^{e}_{i-1},t^{e}_i),
\label{eq:LMMe}
\end{equation}%
where $y^{e}_0 = y^g$ and $y^{e}_{N_e}$ is the ready-made design.

\subsubsection{Image-into-prompt Editing}
\label{sec:IE}
Different from the vanilla method, the input of $N_m$-rounds customized editing is a combination of text editing-prompt $T^{m}$=$\{t_{i}^{m}\}^{N_m}_1$ and image editing-prompt $Y^{m}$=$\{y^{m}_i\}^{N_m}_1$, defined as $\{<t^{m}_i,y^{m}_i>\}^{N_m}_1$. Correspondingly, the edited images are defined as $\hat{Y}^{e}$=$\{\hat{y}^{e}_i\}^{N_m}_1$ and each $\hat{y}^{e}_i$ is generated by an customized editing $\mathit{LMM}_m$:

\begin{equation}
\hat{y}^{e}_i = \mathit{LMM}_{m}(\hat{y}^{e}_{i-1},<t^{m}_i,y^{m}_i>).
\label{eq:LMMm}
\end{equation}%

Each $t^{m}_i$ is modified from the original text-prompt $t^{e}_i$ in Sec.\ref{sec:TE}, where the referring prompt changes to ``Please edit the first image based on the following description and the second reference image'' plus the description and referring image, such as the example in Fig.\ref{fig:method}. It is worth noticing that $N_m$ and $N_e$ can be the same or different.

\subsubsection{Applications}
\label{sec:case}
We analyze three applications in Fig.\ref{fig:method}:

\textbf{Sketch Image:} Sketch images typically derive from hand drawing. Such images contain the core concepts of fashion design but appear relatively rough (\textit{e.g.}, line drawing). As shown in the first iteration, language often struggles to convey the professional design's core concepts, where image references tend to be more precious.

\textbf{Detailed Image:} Detail images are typically needed from fine-grained modification requests, which provide precision changes (\textit{e.g.}, number, color, or layout). As shown in the second iteration, foundation LMMs struggle to capture these details, even is provided text descriptions (\textit{e.g.}, 4 buttons arranged in a specific location).

\textbf{Example Image:} Sample images typically derive from designs by professional designers. Such designs generally include complex details that are difficult to describe from normal people. As shown in the last iteration, the user desires a specific item (\textit{e.g.}, boutonniere) from a real image.


\begin{figure}[t]
  \includegraphics[width=0.8\columnwidth]{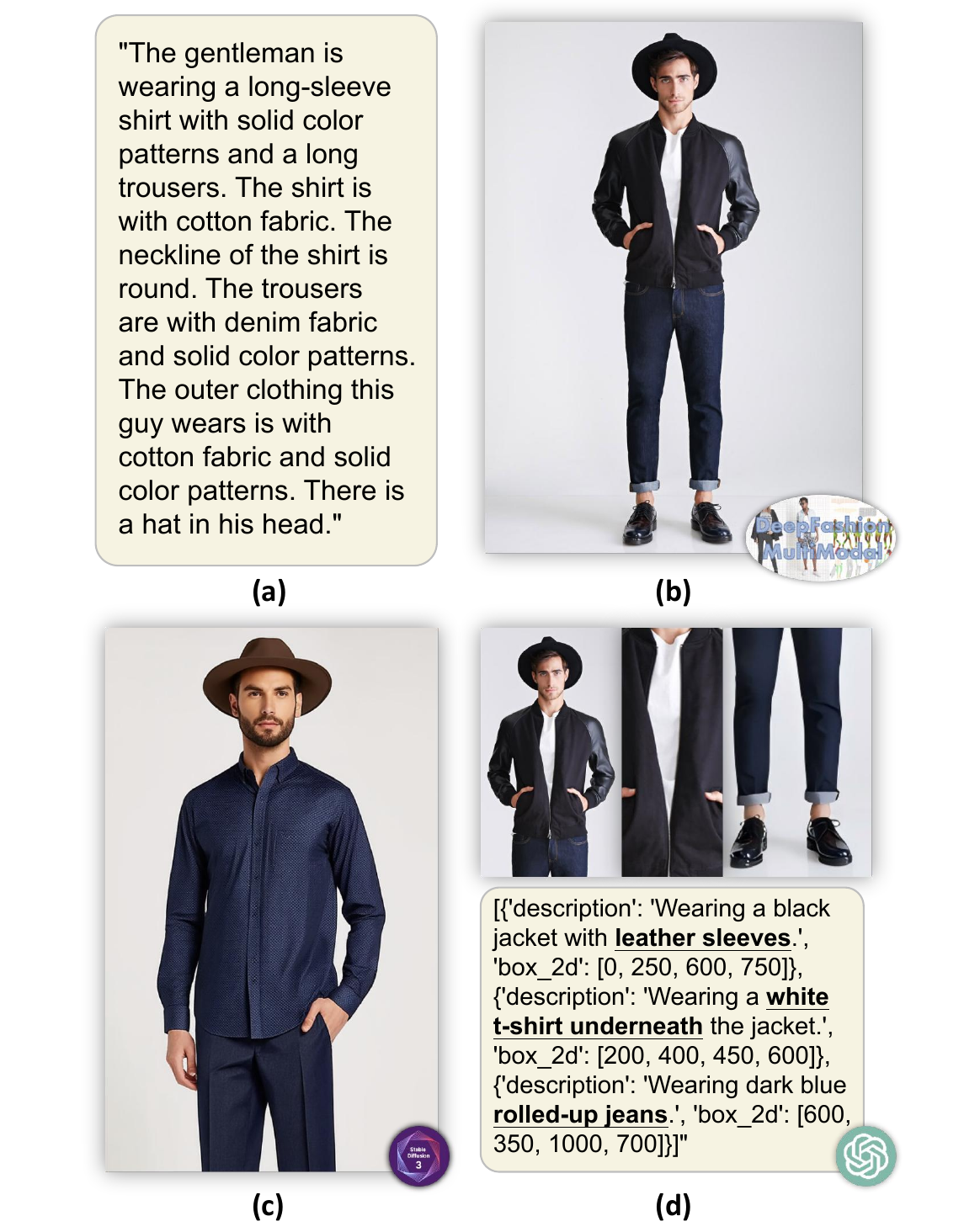}
  \caption{Example of our FashionEdit dataset from original DeepFashion-MultiModal. \textbf{(a)} shows the manual text annotations of the original dataset; \textbf{(b)} presents the original image; \textbf{(c)} is the image generated by SD3~\cite{esser2024scaling} based on (a); \textbf{(d)} contains the image patches, descriptions and location derived from the differences between (b) and (c).}
  \label{fig:dataset}
\end{figure}

\subsection{FashionEdit Dataset}
\subsubsection{Dataset Generation}

DeepFashion-MultiModal~\cite{jiang2022text2human} is a large-scale, high-quality fashion-oriented dataset containing rich multi-modal annotations. It provides human-annotated descriptions with fine-grained labels on two dimensions: clothes colors and clothes fabrics. One example is shown in Fig.\ref{fig:dataset}(a) and (b).

To precisely validate the fine-grained customized control fashion design task, we create a subset from DeepFashion-MultiModal called FashionEdit. Two more components are further developed: generated images and the differences between generated and original images, as shown in Fig.\ref{fig:dataset}(c) and (d). The creation process is as follows:

(1) Images Generation: To simulate the current real-world clothing design processes, we employ the latest SD3~\cite{esser2024scaling} to obtain AI-generated design images from pure descriptions. After the initial generation, we further filtered the top 11,546 images based on CLIP similarity to minimize the noise of the generation process (\emph{e.g.} multiple humans). The train/val separate proportion is 10546/1000 in experience. 

(2) Differences Analysis: To simulate human clothing modification (instruction + image input), we need the descriptions of the change and image parts between the generated and original images. Thus, we implement GPT4.1-mini~\cite{gpt41mini} to analyze the Top-3 differences, and output the description with coordinates from the original images. The prompt is structured as follows:

\begin{lstlisting}[language=Python]
# 1. Task Definition
      Detect the three detailed differences in the clothes between the two images and return a JSON style.
# 2. Problem Definition
      For each result, the description should only contain the difference of the first image, and give the bounding box of the box_2d should be [ymin, xmin, ymax, xmax], normalized to 0-1000.
# 3. Example of output
      The output format is limited to:"[{'description': 'Wearing ...', 'box_2d': [0, 250, 600, 750]}"
\end{lstlisting}

\begin{figure*}[htb]
  \centering
  \includegraphics[width=0.9\linewidth]{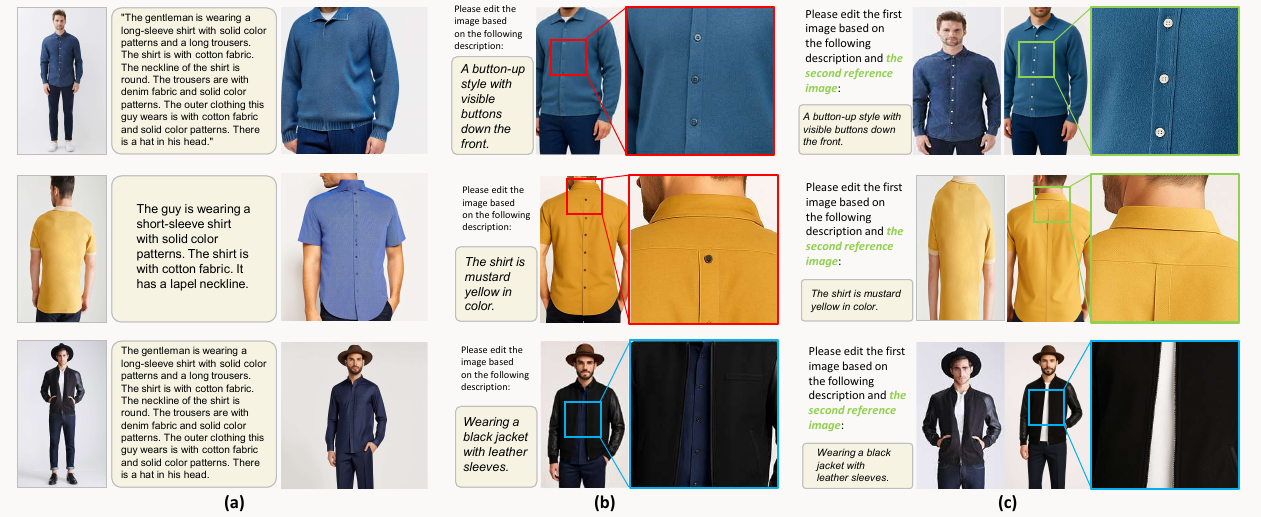}
  \caption{Visualization of the results based on the validation set of FashionEdit. For each example from left to right, \textbf{(a)} combines the ground truth image, ground truth description, and initial generated image; \textbf{(b)} is the result of text-as-prompt (Sec.\ref{sec:TE}) combines the text-prompt, modified image with detail; \textbf{(c)} is the result of image-into-prompt (Sec.\ref{sec:IE}) combines the text-prompt, image-prompt, and modified image with detail.}
  \label{fig:exp}
\end{figure*}

\section{Experiments}
\subsection{Experimental Settings}


To evaluate the generated image, we use the CLIP~\cite{radford2021learning} similarity scores for generation similarity, the CLIP* for user satisfaction, and PSNR for quality. (1) \textbf{CLIP} is computed via cosine distance, which measures the similarity between high-dimensional image embeddings via the CLIP encoder; (2) \textbf{CLIP*} is calculated by counting the number of generated images with a high CLIP score (>90\%), deriving the number of validation images; (3) \textbf{PSNR} is computed by the logarithmic ratio of peak reference signal power to reconstruction error power.

\begin{table}
\centering
\begin{tabular}{lll>{\columncolor{gray!20}}ll}
\hline
\hline
\multirow{1}{*}{Method} & \multirow{1}{*}{Prompt}  & \multicolumn{1}{c}{CLIP \tiny ($\uparrow$)}  & \multirow{1}{*}{CLIP* \tiny ($\uparrow$)} & \multicolumn{1}{c}{PSNR \tiny ($\uparrow$)} \\
 \hline
 \hline
  SD2\tiny(train+val) & text & 69.25 & 0.00  & 6.93 \\
  SD3\tiny(train+val) & text & 82.85 & 1.75 & 9.94 \\
  \hline
  Vanilla\tiny(1) & text & 85.77 & 13.6 & 9.45 \\
  Vanilla\tiny(2) & text & 85.78 & 14.9 & 9.29 \\
  Vanilla\tiny(3) & text & 85.60 & 15.1 & 9.30\\
  \hline
\textbf{BUG}\tiny(1) & text+image  & 87.27 & 26.4 & 9.75 \\
\textbf{BUG}\tiny(2) & text+image & 87.77  & 30.9 & 9.76 \\
\textbf{BUG}\tiny(3) & text+image & \textcolor{blue}{\textbf{87.91}}  & \textcolor{blue}{\textbf{35.4}} &\textcolor{blue}{\textbf{9.96}} \\
\hline
\hline
\end{tabular}
 \caption{Experiences of initialization methods, vanilla methods (different steps), and image-into-prompt methods (different steps) according to the CLIP, CLIP*, and PSNR scores on FashionEdit datasets. $\uparrow$: Higher is better.}
  \label{tab:SOTA}
\end{table}

\subsection{Experimental Results}
\subsubsection{Comparisons with different methods}

Through comparisons of different image generation models (upper Tab.\ref{tab:SOTA}), the latest SD3~\cite{esser2024scaling} outperforms SD2~\cite{rombach2022high} in the entire ``train+val'' sets of FashionEdit, where the most significant improvement is on the CLIP score from 69.25\% to 82.85\%. It is worth noticing that the CLIP* score is low for both SD2 and SD3, which indicates the dissatisfaction of the first generation, proving the necessity for image editing in the fashion design task.

Analyzing the different settings for vanilla methods (central Tab.\ref{tab:SOTA}) with ``(1)'' to ``(3)'' modification steps in the ``val'' set of FashionEdit, CLIP* score continually increases from 13.6\% to 15.1\% with a 1.5\% improvement, which demonstrates the effectiveness of multiple editing. Similar results are also reported for BUG settings with 9.0\% improvement. It is worth noticing that the CLIP* score between ``Vanilla(3)'' and our ``BUG(3)'' after three modifications increased 20.3\%, proving the potential of BUG under multiple modifications.

The best result comes from our BUG method with three modifications (lower Tab.\ref{tab:SOTA}), where the CLIP score reaches 87.91\%, CLIP* score reaches 35.4\%, and PSNR score reaches 9.96\%, respectively. This demonstrates the effectiveness of our image-to-prompt benchmark for a better understanding of generation. (GPT4.1-mini~\cite{gpt41mini})

\subsubsection{Visualization of Different Methods}

In Fig.\ref{fig:exp}, we provide three sets of visualization results to compare the text-as-prompt and image-into-prompt methods (one-reference setting), based on the validation set of our FashionEdit.

Our method provides a more detailed image generation (Case 1). For example, though both methods add visible buttons on shirts following the request, the image-into-prompt result matches the referring cropped image on button color in ``white''. This demonstrates the necessity of referring image, where the text description misses details.

Besides, our method handles the physical conflict to generate a more realistic image (Case 2). For example, the draft design is weird for creating a front-side shirt with a back-side human position. The text-as-prompt result changes the color of the shirt following the instruction, but ignores the conflict, while the BUG result flips the shirt into the right position. This demonstrates the importance of physics laws in the referring images.


Our method prioritizes the referring image over the text description (Case 3). For example, the image-into-prompt result corrects the underlayer of ``cotton white shirt'', while the text editing only adds the missing ``jacket'' based on the original wrong T-shirt. Our method enhances the design workflow where the user can choose from multiple outputs.

\section{Conclusion}

Our work analyzes the core challenge in fashion design with GenAI, and proposes a BUG benchmark that adopts both text-prompts and image-prompts under iterative image editing. Experience proves that our benchmark dramatically improves user satisfaction under the following instructions. The new FashionEdit dataset, which simulates the real-world clothing design workflow, provides a new possibility to further employ AI in the arts industry.


\bibliographystyle{IEEEbib}
\bibliography{strings,refs}

\end{document}